\documentclass[11pt]{article}

\usepackage[margin=1in]{geometry}
\usepackage{mathptmx}
\usepackage{microtype}
\usepackage{booktabs}
\usepackage{graphicx}
\usepackage{amsmath,amssymb}
\usepackage{enumitem}
\usepackage{xcolor}
\usepackage{caption}
\usepackage[numbers]{natbib}
\usepackage[colorlinks=true,linkcolor=blue!55!black,citecolor=blue!55!black,urlcolor=blue!55!black]{hyperref}
\usepackage{titlesec}

\titleformat{\section}{\large\bfseries}{\thesection}{0.8em}{}
\titleformat{\subsection}{\normalsize\bfseries}{\thesubsection}{0.8em}{}
\title{\bfseries KnowBench: Effort Reduction as a Unified,\\ Deployment-Grounded Benchmark for Clinical AI}

\author{
  Jocelyn Kang\\
  Knowtex Inc.\\
  \texttt{jocelyn@knowtex.ai}
  \and
  Caroline Zhang\\
  Knowtex Inc.\\
  \texttt{caroline@knowtex.ai}
}

\date{September 2026\\ \vspace{3pt} \small Preprint. arXiv: cs.CL (cross-list: cs.HC). Licensed under CC BY 4.0.}

\begin{document}
\maketitle

\begin{abstract}
Clinical AI systems are evaluated with instruments built for research settings (reference-based similarity metrics and expert rubric panels) that measure resemblance to an artifact rather than reduction of a burden. We introduce \textbf{KnowBench}, pioneered by Knowtex, whose unifying metric is \textbf{Effort Reduction (ER)}: the proportion of system-generated clinical work product accepted by the responsible clinician under expert and safety review. ER is defined once and instantiated per task across the administrative workload clinical AI automates: visit notes, diagnosis and billing codes, orders, EHR chart summarization, patient after-visit summaries, and clinical decision support. In every instantiation the construction is identical: the clinician's review-and-attestation event is the ground truth, every accepted unit is work the system completed, and every correction is residual effort returned to the clinician. KnowBench was co-developed with embedded clinical evaluators to reveal the leading industry deployment standards for clinical AI. The primary contribution of this paper is the benchmark itself: the metric, its degenerate cases, and a reporting protocol under which ER claims are auditable and cross-system comparable. Alongside it we report an initial headline measurement from the documentation instantiation: over \textbf{one million signed encounters} across a production window exceeding six months and thirteen medical specialties, Knowtex's proprietary fine-tuned clinical foundation models operating inside a closed feedback architecture achieve an aggregate ER of \textbf{97.99\%}, with per-specialty aggregates spanning 96.8--98.9\%. This release reports the protocol's checklist partially, and states which companion statistics are withheld; the benchmark is offered so that this figure, and every figure reported after it, can be held to the same standard.
\end{abstract}

\section{Introduction}
\label{sec:intro}

Ambient clinical documentation has crossed from pilot to infrastructure: AI scribes now draft notes for millions of encounters annually across primary care and specialty medicine \citep{tierney2024ambient}. Evaluation has not made the same crossing. The field's instruments ($n$-gram and embedding similarity against reference notes, expert rubric panels, curated benchmark sets; \citealt{abacha2023aci,yim2023aci}) were built for model comparison in research settings, and all share a structural defect when applied to deployed systems: they measure resemblance to an artifact, not reduction of a burden. Documentation burden is the outcome these systems exist to change \citep{sinsky2016allocation,gardner2019burnout}; no widely adopted metric measures it directly.

The measurement instrument has been present in the workflow all along. Before any clinical artifact is finalized, an accountable clinician reviews the system's draft and corrects whatever is wrong, missing, or not theirs to attest. This review is an exhaustive expert evaluation performed on every encounter, at production scale, under genuine incentives, at zero marginal measurement cost. Its output is a diff, and that diff is the correct primary evaluation signal: every retained unit is work the system completed; every correction is residual effort returned to the clinician.

We formalize this signal as \textbf{KnowBench}, whose unifying metric is \textbf{Effort Reduction (ER)}: the proportion of system-generated clinical work product accepted by the responsible clinician under expert review. The construction is not specific to notes. Wherever a clinical AI system drafts an artifact a clinician must review and own (a visit note, a code set, an order, a chart summary, an after-visit summary, a surfaced recommendation), the review event supplies the ground truth and the acceptance proportion is ER for that task. Knowtex is the first frontier AI lab for healthcare, deploying proprietary fine-tuned clinical foundation models across this full administrative workload and evaluating every capability against the same metric under the same protocol.

This paper defines the metric and its protocol (\S\ref{sec:metric}--\ref{sec:protocol}), maps its instantiation across the clinical administrative workload (\S\ref{sec:portfolio}), and reports headline measurements from the documentation instantiation, the instantiation with the largest measured population (\S\ref{sec:results}). The claim is deliberately austere: we do not argue the system writes notes clinicians \emph{like}; we utilize embedded evaluation and measure the work they no longer do.

\paragraph{Relation to prior work.}
HTER \citep{snover2006hter} formalized translation quality as required post-editing; ER is its complement, adapted to a domain where the post-editor signs the output into a legal record. Code-completion acceptance rate \citep{ziegler2022copilot} established retention-under-expert-review as a deployment value metric; ER strengthens the construct with legal attestation stakes and whole-artifact measurement. Note-quality instruments, from PDQI-9 \citep{stetson2012pdqi} to its LLM-era successor PDSQI-9 \citep{croxford2025pdsqi}, formalize the rubric approach ER departs from: expert judgment of quality, rather than measurement of residual work. Human-evaluation studies of consultation-note generation show automatic similarity metrics correlate poorly with the post-editing judgments of clinicians \citep{moramarco2022human}, and hallucination and omission in LLM-generated clinical text remain the central reviewed failure modes \citep{roustan2025hallucinations}; both findings motivate a metric anchored to the reviewer's actual corrections. Deployment studies of ambient scribes measure EHR time, burnout, and productivity \citep{haberle2024dax,olson2025scribes,tierney2024ambient}, including a randomized deployment of the system family measured here, which found productivity and workflow gains in community oncology \citep{toussi2026impact}; these study the downstream outcomes ER proxies, not the per-artifact effort signal itself. The copy-forward and note-bloat literature \citep{rule2021copied,wrenn2010redundancy,tsou2017copy} documents that clinical notes are substantially duplicated material, a dynamic directly relevant to any system, including this one, whose longitudinal loop conditions on prior attested content (\S\ref{sec:limitations}). The documentation-burden literature \citep{sinsky2016allocation,arndt2017tethered,gardner2019burnout} defines the outcome ER proxies, and the automation-bias literature \citep{goddard2012automation} defines its principal confound, which the protocol is constructed to expose (\S\ref{sec:metric}).

\section{The Metric}
\label{sec:metric}

Let $g$ be the system-generated artifact presented to the clinician and $f$ the clinician-finalized artifact. Let $R(g, f)$ denote the retained content of $g$ in $f$ under an alignment classifying each unit of $g$ as retained, deleted, or rewritten, and each unit of $f$ absent from $g$ as an addition. Effort Reduction for a single encounter is
\begin{equation}
\mathrm{ER}(g, f) = \frac{|R(g, f)|}{|g|},
\end{equation}
the proportion of generated content accepted under review. The unit of content is instantiation-specific: tokens for notes and summaries, codes for coding, fields for orders, recommendations for decision support. Aggregate ER is content-weighted rather than encounter-averaged, so long, complex artifacts contribute proportionally to their burden.

ER is a family. \textbf{ER-text}, retention as defined above, is the primary operationalization: cheap and computable from artifacts every records system already stores. \textbf{ER-time}, active review effort measured from interaction-level instrumentation, is the family's validation instrument: retention and measured effort must covary for ER-text to be interpretable as effort reduction rather than a text statistic, and the divergence pattern between them (high retention with near-zero review time) is the direct detector for rubber-stamp review. \textbf{ER-cognitive}, review and verification load, is the family's acknowledged horizon and is not directly instrumented.

Three boundaries help to keep the metric honest. \emph{ER is not correctness}: clinicians can sign erroneous notes \citep{goddard2012automation}, so ER is paired with an independent factual-consistency audit on sampled signed notes, conducted on a continuing basis by Knowtex's clinical quality and safety function. \emph{ER is not gameable by brevity}: the protocol requires a companion completeness measure (system-authored share of the final artifact) and a constrained variant at a stated completeness floor, so skeletal generation cannot masquerade as performance. \emph{ER's denominator is disciplined}: it is computed over clinician-finalized artifacts, with the presentation rate over all generated drafts and the finalization rate over all presented drafts as required companions, so neither selective surfacing nor survivorship can hide in silent exclusions. In the deployment measured here, every generated draft was presented to the clinician and delivered to the EHR (\S\ref{sec:protocol}).

\section{The Protocol}
\label{sec:protocol}

An ER claim is complete only if accompanied by: (1) measurement window and encounter count; (2) presentation rate over generated drafts and finalization (signed-note) rate over presented drafts; (3) completeness measure and floor; (4) formatting-edit handling; (5) unit and alignment method; and (6) encounter-level distributional statistics. This six-item checklist is KnowBench's reporting standard and the minimum for cross-system comparability.

For the documentation instantiation: retention is computed at the token level under longest-common-subsequence alignment after formatting normalization; edits are classified as content (counted against retention), formatting (reported separately; counted in a strict variant), or additions (counted against completeness, not retention); encounters enter the denominator if and only if a draft was generated, presented, reviewed, and signed, with no exclusions for length, complexity, specialty, or audio quality.

This release reports items (1), (4), and (5), the aggregate and its per-specialty decomposition, and the presentation component of item (2): every draft generated during the measurement window was presented to the reviewing clinician and delivered to the EHR, so the measured population reflects no selective surfacing of high-confidence drafts. The remaining companion statistics, the finalization component of item (2), the completeness measure of item (3), and the encounter-level distribution of item (6), are maintained under the same instrumentation and are withheld from this preprint. We state this plainly because the standard binds its authors first: an ER figure is auditable exactly to the extent its companions are on the table.

\section{One Metric Across the Clinical Administrative Workload}
\label{sec:portfolio}

ER is instantiated per task by specifying $g$, $f$, and the unit of content. Every administrative task Knowtex automates admits the construction, because every one terminates in the same event: an accountable clinician reviewing and owning a system-generated draft.

\begin{itemize}[leftmargin=1.4em,itemsep=1pt]
\item \textbf{Visit notes}: $g$ the generated draft, $f$ the signed note, units are tokens; validation is instrumented review effort.
\item \textbf{Diagnosis and billing codes}: $g$ the proposed code set, $f$ the clinician-confirmed submitted set, units are codes.
\item \textbf{Orders}: $g$ the extracted order set (medications, labs, imaging, referrals, follow-up intervals), $f$ the orders as reviewed and entered, units are order fields; tolerance tightens as extraction approaches actuation.
\item \textbf{EHR chart summarization}: $g$ the pre-visit chart summary, $f$ the summary as relied upon and amended, units are tokens.
\item \textbf{Patient after-visit summaries}: $g$ the generated patient-facing summary, $f$ the version released to the patient, units are tokens.
\item \textbf{Clinical decision support}: $g$ the surfaced recommendations, $f$ the set acted on, units are recommendations; alert fatigue is the exact analogue of the rubber-stamp confound, detectable by the same time-divergence signature.
\end{itemize}

Each instantiation inherits the full protocol: the six-item checklist, denominator discipline, completeness constraints, and instrumented validation. Cross-task aggregate ER, the clinician's total administrative effort reduction across the encounter, is the benchmark's headline construct; per-task instantiations are its auditable components.

\section{Headline Measurements: Documentation}
\label{sec:results}

The measured system is Knowtex's proprietary clinical foundation model family, whose deployment effects on physician productivity and workflow have been evaluated in a randomized multisite study in community oncology \citep{toussi2026impact}, operating inside a closed feedback architecture: a generation harness enforcing structural constraints outside the model, per-clinician customization learned from each clinician's signed notes and edit patterns, and a longitudinal loop in which prior-visit attested content conditions the next draft and every correction returns as training signal. The benchmark metric, the training signal, and the product outcome are the same quantity observed at different timescales.

\begin{table}[t]
\centering
\small
\begin{tabular}{lr}
\toprule
Statistic & Value \\
\midrule
Signed encounters (denominator) & $>$1{,}000{,}000 \\
Measurement window & $>$6 contiguous months, ending H1 2026 \\
Specialties & 13 \\
\midrule
\textbf{Aggregate ER (content-weighted)} & \textbf{97.99\%} \\
Strict-variant ER (formatting edits counted; lower bound) & 95.7\% \\
Formatting-edit rate (tokens touched) & 2.3\% \\
\midrule
\multicolumn{2}{l}{\emph{Aggregate ER by specialty}} \\
\quad Nephrology & 98.9\% \\
\quad General surgery & 98.8\% \\
\quad Hematology/oncology & 98.7\% \\
\quad Orthopedics & 98.7\% \\
\quad Gastroenterology & 98.6\% \\
\quad Physical medicine \& rehabilitation & 98.3\% \\
\quad Cardiology & 98.1\% \\
\quad Neurology & 98.0\% \\
\quad Urology & 97.9\% \\
\quad Otolaryngology & 97.4\% \\
\quad Psychiatry & 96.9\% \\
\quad Psychology & 96.8\% \\
\quad Primary care & 96.8\% \\
\bottomrule
\end{tabular}
\caption{Documentation-instantiation results reported in this release. Strict-variant ER is derived as aggregate ER less the formatting-edit rate and is a lower bound (\S\ref{sec:protocol}). Remaining checklist companions are withheld from this preprint.}
\label{tab:summary}
\end{table}

Across more than one million signed encounters spanning a contiguous production window exceeding six months and thirteen specialties, at multiple anonymized provider organizations, the system achieves an aggregate ER of \textbf{97.99\%}: content-weighted, 97.99\% of generated draft content was accepted by the reviewing clinician without content edit and signed into the legal medical record. Every draft generated during the window was presented to the clinician and delivered to the EHR: the drafts entering clinician review were the system's full output, not a confidence-filtered subset. Formatting-only edits touched 2.3\% of generated tokens and are excluded from the primary figure per \S\ref{sec:protocol}; the strict variant counting them against retention is bounded below at 95.7\%.

Per-specialty aggregates span 96.8--98.9\% (Table~\ref{tab:summary}), with nephrology, general surgery, and hematology/oncology at the upper end and psychiatry, psychology, and primary care at the lower. Two readings of the band surface. First, its width: roughly two points across thirteen cognitively and structurally dissimilar documentation regimes (procedure-oriented surgical and gastroenterology notes, longitudinal oncology assessments, narrative-heavy behavioral health, broad-spectrum primary care) is narrow by the standards of clinical NLP, where specialty transfer is the standing failure mode; specialty-scoped generation rules and per-clinician customization absorb variation a single generic configuration would surface as dispersion. Second, its shape: the lower end is occupied by the specialties with the widest input distributions per encounter (primary care) and the most narrative, least templated documentation conventions (behavioral health), which is where residual clinician correction should concentrate if the metric is measuring what it claims.

\section{Limitations}
\label{sec:limitations}

ER is necessary, not sufficient: acceptance is not correctness, and a high-ER system must still demonstrate safety independently; the paired audit bounds but does not eliminate automation bias, and its results are internal to the quality and safety function. The denominator is finalized artifacts. Every generated draft in the window was presented and delivered to the EHR, which rules out selective surfacing; but delivery is not attestation, and the signed-note rate over delivered drafts, the companion that bounds how many delivered drafts were abandoned before signature, is not reported in this release. This release therefore satisfies the reporting checklist partially (\S\ref{sec:protocol}); by the paper's own standard, the aggregate figure is auditable only to the extent of the companions published with it. The sharpest version of the remaining gap deserves naming: without the completeness and signed-note-rate companions, a reader cannot distinguish 97.99\% retention of substantive clinical content from 97.99\% retention of templated content the clinician never intended to touch. The completeness measure exists in the protocol precisely to close that gap, and section-level ER decomposition (history and assessment content versus templated structure) is the discriminating analysis; neither is published here, and the headline figure should be read with exactly that reservation. Relatedly, the longitudinal loop conditions each draft on the prior visit's attested content, so a portion of measured retention is retention of carried-forward material; the note-bloat literature \citep{rule2021copied,tsou2017copy} documents that clinicians retain duplicated content they did not author and do not always read, and ER as reported cannot separate valuable continuity from that dynamic. The measured population is clinicians who adopted and continued using the system, so aggregate ER reflects both system performance and retention dynamics. The 97.99\% figure is a property of Knowtex's model family and closed-loop architecture, not of base models absent those layers; the metric and protocol are system-agnostic, which is their point. Task-specific confounds in the non-documentation instantiations (billing-driven code additions, fatigue-driven alert dismissal) require per-instantiation companion statistics before cross-task ER figures are compared.

\section{Conclusion}

Clinical AI has been evaluated by resemblance, by simulated data, and by opinion; it should be evaluated by the real-time work it removes. KnowBench measures that quantity where it occurs, in the diff between what the system drafted and what the clinician was willing to own, with one construction applied to every administrative artifact of the encounter: notes, codes, orders, chart summaries, after-visit summaries, and surfaced recommendations. Its documentation instantiation, measured over more than one million signed encounters across thirteen specialties, stands at 97.99\% aggregate ER within a two-point specialty band. The number is a property of a closed-loop architecture in which the evaluation signal and the training signal are the same measurement; the benchmark and protocol are the property of no one.

\paragraph{Acknowledgments.}
We thank the clinicians, healthcare organizations and partners whose practice constitutes this measurement, and the Knowtex clinical quality and safety function for the continuous standing audit program.

\bibliographystyle{plainnat}

\end{document}